\documentclass{optica-article}
\usepackage{ulem}
\journal{opticajournal} 

\articletype{Research Article}
\usepackage{doi}

\begin{document}

\title{Single-pixel 3D imaging based on fusion temporal data of single photon detector and millimeter-wave radar}

\author{Tingqin Lai,\authormark{1,2} Xiaolin Liang,\authormark{1,2} Yi Zhu,\authormark{3} Xinyi Wu,\authormark{1,2} Lianye Liao,\authormark{1,2} Xuelin Yuan,\authormark{1,2} Ping Su,\authormark{3} and Shihai Sun\authormark{1,2,*}}

\address{\authormark{1}School of Electronics and Communication Engineering, Sun Yat-sen University, Shenzhen, Guangdong 518107, P.R. China.\\
\authormark{2}Shenzhen Campus of Sun Yat-sen University, No. 66, Gongchang Road, Guangming District, Shenzhen, Guangdong 518107, P.R. China\\
\authormark{3}Tsinghua Shenzhen International Graduate School, Tsinghua University, Shenzhen, 518055\\
}

\email{\authormark{*}sunshh8@mail.sysu.edu.cn} 


\begin{abstract*} 
Recently, there has been increased attention towards 3D imaging using single-pixel single-photon detection (also known as temporal data) due to its potential advantages in terms of cost and power efficiency. However, to eliminate the symmetry blur in the reconstructed images, a fixed background is required. This paper proposes a fusion-data-based 3D imaging method that utilizes a single-pixel single-photon detector and a millimeter-wave radar to capture temporal histograms of a scene from multiple perspectives. Subsequently, the 3D information can be reconstructed from the one-dimensional fusion temporal data by using Artificial Neural Network (ANN). Both the simulation and experimental results demonstrate that our fusion method effectively eliminates symmetry blur and improves the quality of the reconstructed images.
\end{abstract*}

\section{Introduction}

%
%

3D imaging finds wide applications in systems such as automatic driving target recognition, Unmanned Aerial Vehicle (UAV) automatic navigation, and reconnaissance surveillance. Three main methods are commonly used for 3D imaging: structured light\cite{jiegouguang1,jiegouguang2,jiegouguang3}, binocular stereo imaging\cite{shuangmu1,shuangmu2,shuangmu3,shuangmu4}, and Time-of-Flight (ToF)\cite{tof1,tof2,tof3}. ToF imaging has gained significant attention due to its advantages, including medium and long-distance measurement capability, high precision, and strong anti-interference ability. Direct Time-of-Flight (D-ToF) methods estimate depth information by measuring the flight time of light from the scene to the sensor. 
Currently, three mainstream methods are used to capture lateral spatial information in ToF imaging: SPAD (Single-Photon Avalanche Diode) array sensor\cite{SPAD1,SPAD2,SPAD3,SPAD4}, single-pixel imaging combined with specialized structured lighting\cite{ghost1,ghost3,ghost4,DMD,DMO1,DMO2}, and single-pixel imaging combined with scanning techniques\cite{scan1,scan2,scan3,KILO,xu1,xu2,xu3}. Due to the high cost of SPAD array sensors, the low-cost and highly sensitive single-pixel imaging method has gained more attention. In recent years, single-pixel imaging has also produced significant research results. 
This method utilizes low-cost color LED arrays for structured illumination, enabling color imaging with single photodiodes\cite{LED1}. Moreover, by adding a small number of photodiodes at different positions, the method can be extended to three-dimensional imaging.
While the above method combines multiple single-pixel detectors for three-dimensional imaging, it does not achieve true single-pixel three-dimensional imaging. Ghost imaging employs a computer-controlled Spatial Light Modulator (SLM) to generate speckle pattern illumination on the object, eliminating the requirement for beam splitters and array detectors. Imaging is accomplished by synchronously measuring the light intensity from the bucket detector\cite{ghost1,ghost2,ghost3,ghost4}. 
Otherwise, a pulsed laser uniformly illuminates a Digital Micromirror Device (DMD), used to provide structured illumination onto a scene, and the back-scattered light is collected onto a photodiode. The measured light intensities are used in a 3D reconstruction algorithm to reconstruct both depth and reflectivity images\cite{DMD,DMO1,DMO2}. However, specialized structured lighting is typically not suitable for long-distance 3D imaging. In contrast, combining single-pixel SPAD with scanning structures enables long-distance 3D imaging with high spatial resolution\cite{scan1,scan2,scan3,KILO}. By implementing time filtering to suppress noise, single-pixel imaging can achieve an average imaging sensitivity of 0.4 signal photons per pixel, enabling long-range three-dimensional imaging up to 200 km\cite{xu1,xu2,xu3}.

To acquire lateral spatial information of targets, the aforementioned single-pixel imaging methods rely on array sensors, specialized structured lighting, or single-pixel scanning techniques. However, the high costs and complex structures associated with these approaches hinder the progress of single-pixel imaging. Consequently, alternative methods that do not involve scanning or unique lighting structures have been proposed in recent years\cite{temprol1,temprol2,temprol3}. These methods, known as 3D imaging from temporal data, capture the temporal information (time-of-flight) of the entire scene using a single-pixel Single-Photon Detector (SPD) and a Time Digital Converter (TDC), followed by reconstruction of the 3D images using an ANN\cite{ann1,ann2,ann3,ann4}. However, the aforementioned methods inherently exhibit symmetry blur issues. This arises from the fact that a scene with center symmetry, captured by a single-pixel detector placed at its center, yields identical measurement results. For instance, the same measurement value is obtained for symmetrical positions on the left and right sides of a single-pixel sensor. One approach\cite{temprol1} to address this problem is to introduce a background that reveals the relative position of the subject. Another strategy\cite{temprol3} involves leveraging multipath time signals to gather more scene-related data. However, both of these methods heavily rely on the specific requirements of the scene and may fail to produce accurate images if the background is a plain wall or lacks distinguishing features.

This paper proposes a fusion-data-based method for 3D imaging. Our approach involves placing a single-pixel SPD and a millimeter-wave radar at different locations, forming a specific angle with respect to the target. The SPD records the arrival time of return photons from the entire scene as a temporal histogram, while the millimeter-wave radar captures the one-dimensional range profile of the scene. The data from the SPD and radar are directly fused and input into an ANN for 3D scene reconstruction. Millimeter-wave radar is a radar system that operates in the millimeter-wave frequency band (approximately 30-300 GHz). This radar technology has high resolution and accuracy, making it valuable in various applications. In fact, millimeter-wave radar is widely used in many application, such as autonomous vehicles, drones, aviation radar, an so on\cite{mm1,mm2,mm3}. The fusion approach of SPD + millimeter-wave radar is used in this paper, because millimeter-wave radar offers numerous advantages that SPD do not possess. For instance, it has a strong adaptability to adverse weather conditions and can operate normally in rain, smog, and heavy snow. Additionally, millimeter-wave radar is less susceptible to interference from other light sources, such as sunlight or car headlights. More importantly, millimeter-wave radar is generally cheaper and has a higher level of integration. Typically, an entire millimeter-wave radar system can be integrated onto a small circuit board. The added cost of using a system with SPD + millimeter-wave radar, compared to one using only a single-pixel SPD, is negligible. Therefore, we believe that the system architecture of SPD + millimeter-wave radar can fully capitalize on the strengths of both types of sensors and overcome the weaknesses of using a single sensor, achieving a synergistic effect where the whole is greater than the sum of its parts.By integrating the single-pixel SPD and millimeter-wave radar, our method achieves higher accuracy in 3D scene reconstruction and effectively addresses the symmetry blur issue without the need for a background. Both simulation and experimental results demonstrate the successful elimination of symmetry blur and significant improvement in the quality of reconstructed images.

The remainder of this paper is structured as follows: Section II presents a theoretical analysis of the imaging method, addressing the associated challenges, and introduces the imaging algorithm based on ANN. In Section III, numerical simulations are conducted to compare the performance of the two imaging methods under various imaging conditions, demonstrating the effectiveness of the proposed approach. Section IV describes the imaging experiments performed using an optical system operating at 1550 nm and a 60 GHz millimeter-wave radar, assessing the feasibility of the proposed method. Section V discusses the potential capabilities of our method in challenging environments. Finally, Section VI concludes the paper.

\section{System framework and imaging principle}

\begin{figure*}[htbp]
	\centering\includegraphics[width=11cm]{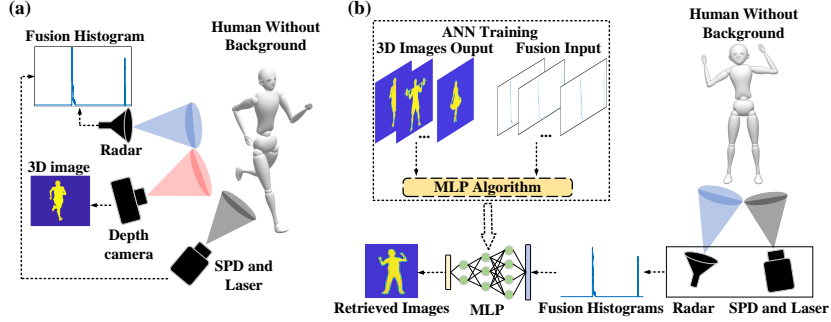}
	\caption{3D imaging with fusion data. (a) Data acquisition process. (b) 3D images recovery process. In 3D images recovery process, the ANN training is performed only once, then the MLP algorithm can directly reconstructs the 3D image from the temporal histogram. The Radar represents the millimeter-wave radar. The human moves in an empty room}
	\label{fig1}
\end{figure*}

The proposed fusion-data-based 3D imaging method is illustrated in Fig.\ref{fig1}, comprising two main processes: data acquisition (Fig.\ref{fig1}(a)) and 3D image recovery (Fig.\ref{fig1}(b)). During the data acquisition process, a millimeter-wave radar is positioned in proximity to a single-pixel SPD at a specific angle relative to the target under measurement. This arrangement mimics binocular imaging as the millimeter-wave radar and SPD are separated by a certain distance. The temporal histogram of the SPD and millimeter-wave data is fused to generate a fusion histogram by connecting. Concurrently, a high-precision depth camera captures the 3D image of the target solely for training purposes, not participating in the image recovery process. By varying the target's orientation and position during data acquisition, a substantial amount of real measured training data is obtained. Each pair of data for 3D image recovery consists of one fusion histogram and one depth map. In the 3D image recovery process, the acquired fusion histogram serves as input, while the depth map functions as output for training an ANN. Sufficient training data and iterations enable the ANN to effectively learn the mapping between the input fusion histogram and depth map. Once trained, the network can directly retrieve new, untrained targets by inputting the acquired fusion histogram data from the millimeter-wave radar and the SPD. 

To describe this method mathematically, it is simple to construct a forward model where all points in the scene that are at some distance, $r_{i}=(x_{i},y_{i},z_{i})$, from the SPD provide a related photon arrival time, $t_{i}=c^{-1}|r_{i}|=c^{-1}\sqrt{(x_{i}^{2}+y_{i}^{2}+z_{i}^{2})}$(where $c$ is the speed of light). By recording the number of photons arriving at different times $t_{i}$, we can build up a temporal histogram $H_{s}$ that contains information about the scene in 3D. This calculation process could be represented as $H_{s} = F_{s}(I)$, where $I$ represents the model of the 3D scene and $F_{s}$ represents the calculation from 3D scene to 1D temporal histogram.

The same purpose as SPD, millimeter wave radar also converts echo collection containing 3D scene information into a 1D histogram. Assume that the millimeter wave radar transmits an Linear Frequency Modulation (LFM) pulse, which is generally in the form of

\begin{equation}
	\label{m1}
	s(t) = w(t)exp[j\phi(t)]
\end{equation}

where $w(t)$ is the real envelop of signal, $\phi(t)$ is the signal modulation phase, $j$ represents the
imaginary unit. For standard LFM signals

\begin{equation}
	\label{m2}
	w(t) = rect(\frac{t}{T_{p}}), \phi(t) = 2\pi f_{c}t+\pi Kt^{2}
\end{equation}

where $rect(x)$ denotes the rectangular function, $T_{p}$ is the pulse width of the signal, $f_{c}$ is the center frequency and $K$ is the chirp-rate. When the distance of the target is $R$, after the LFM pulse reflected by the target, the time delay
of the echo is $t_{0} = 2R/c$, then the echo signal can be expressed as

\begin{equation}
	\label{m3}
	s_{r}(t) = s(t-t_{0}) = rect(\frac{t-t_{0}}{T_{p}})exp[j\pi K(t-t_{0})^{2}]
\end{equation}

To generate the range information of the targets, a matched filter processing method can be performed. Finally, the output is approximately expressed as

\begin{equation}
	\label{m4}
	s_{out}(t) \approx T_{p}sin c[K(t-t_{0})]
\end{equation}

where $sin c (x) = sin (\pi x)/ (\pi x)$. Thus, the target distance information $s_{out}(t)$, which can be represented as 1D histogram $H_{m}$, can be extracted from the echo signal. The process of millimeter wave radar obtaining 1D histogram from 3D scene can also be expressed as $H_{m} = F_{m}(I)$. When the data from SPD and millimeter wave radar are fused, the overall forward model can be expressed as $H = F(I)$.

The goal is to find a mapping $F^{-1}$ that can recover the 3D image $I$ from $H$. The specific methods is to adopt a supervised training approach by gathering a set of temporal histograms corresponding to various scenarios along with the associated ground-truth 3D images taken with a depth camera. These data are then used to train the ANN to find an approximate solution to the $F^{-1}$. When the ANN training is complete (which happens only once), it can reconstruct new 3D images from the temporal histogram that has not been trained. 

\section{Numerical results}
In this section, we conduct simulations of our proposed fusion-data-based 3D imaging model and compare it with the method that utilizes only a single pixel SPD. The specific simulation scenario is depicted in Fig.\ref{fig1}. We position various virtual human models in space and assume that data collection is performed by the millimeter-wave radar located on the left side of the single-pixel SPD at a distance of 0.5 m. Meanwhile, the depth camera is positioned in the center. Simulations are conducted for two cases: with and without background. In the background-free scenario, we exclude the background and solely retain the data corresponding to the human model for simulation. Conversely, in the scenario with background, we situate the human model within a virtual indoor scene for simulation (further details can be found in the supplemental document).

\begin{figure}[htbp]
	\centering\includegraphics[width=11cm]{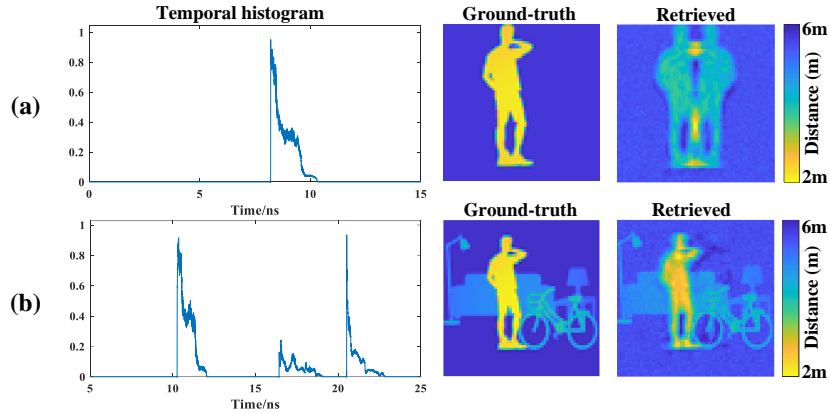}
	\caption{Single-pixel single-photon detector imaging simulation result. (a) Single-pixel 3D imaging symmetry blur result due to lack of background information, (b) 3D imaging without symmetry blur in the background}
	\label{fig2}
\end{figure}

To assess the potential performance under ideal conditions, which represent the best capabilities of current equipment, we conducted simulations with a system time resolution (impulse response function IFR) set to 2 ps (further details regarding the analysis of imaging resolution can be found in the supplemental document). When utilizing only the temporal histogram from a single-pixel SPD for training, the resulting image after passing through the ANN closely resembles the one illustrated in Fig.\ref{fig2}(a).
Consequently, if a single-pixel SPD alone is employed to detect a target in the absence of background, the MLP model fails to differentiate between the left and right sides of the detector, resulting in an image with superimposed left and right targets. Currently, the prevalent solution for single-photon detectors involves incorporating a background behind the subject, as depicted in Fig.\ref{fig2}(b). By including asymmetric targets within the background, symmetry blur is eliminated, and the background imparts relative positional information during the neural network training of the scene (Turpin et al.\cite{temprol1}). An alternative approach entails utilizing the multipath effect of radio frequency or acoustic waves within a confined space to enhance the amount of information and eliminate symmetry blur (Turpin et al.\cite{temprol3}).

\begin{figure*}[bp]
	\centering\includegraphics[width=13cm]{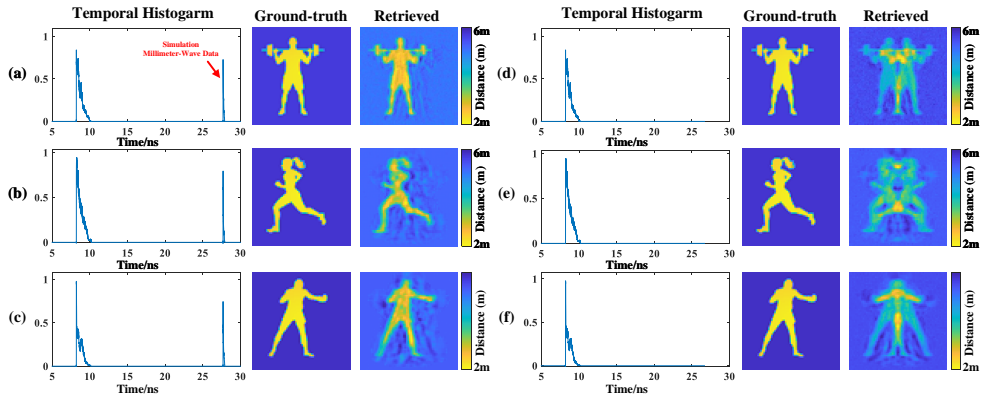}
	\caption{Simulation and reconstruction results of the fusion method. (a)-(c) Images recovered using a fused data-based 3D image reconstruction algorithm. (d)-(f) Images recovered using only single photon data. The first column shows fused temporal histograms generated by simulation [rows (a)–(c)] or histograms with only single-photon data [rows (d)–(f)], The second column shows the ground truth depth maps generated by simulation, and the third column shows the reconstructed images of the time histogram recovered by  the MLP algorithm}
	\label{fig3}
\end{figure*}

In real-world scenarios, certain locations such as flat ground, long corridors, and open rooms may lack sufficiently complex backgrounds and multipath signals to provide relative positional information in the data. This gives rise to the inevitable problem of symmetry blur when conducting target detection in such environments. When the scene is limited and unchangeable, addressing the challenge of symmetry blur requires improvements in the detection-end sensor. In this regard, we propose a cost-effective solution that integrates a millimeter-wave radar sensor and introduces a fusion-data-based 3D imaging method.

As depicted in Fig.\ref{fig3}(a)-(c), the algorithm that fuses millimeter-wave data successfully eliminates symmetry blur and achieves clear reconstruction of the human body. Conversely, Fig.\ref{fig3}(d)-(f) demonstrate that the single-pixel single-photon method produces symmetry blur in the absence of background, rendering it impossible to discern the specific location of the target.

\section{Experimental results}
\begin{figure*}[htbp]
	\centering\includegraphics[width=12cm]{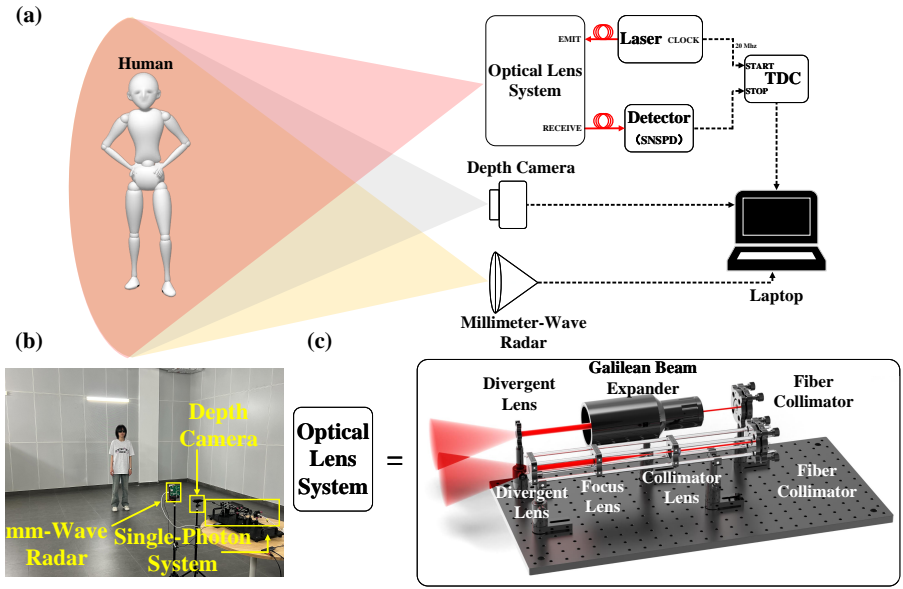}
	\caption{Schematic diagram of the experimental system and experimental scene. (a) Schematic of the layout of the 1550nm fusion-data-based single-photon single-pixel 3D imaging system which comprises a supercontinuum laser source, an SNSNP, a TDC module, and Optical lens system, a depth camera, a millimeter-wave radar, and a laptop. (b) experimental scene with a person. (c) Schematic diagram of an optical lens system}
	\label{fig4}
\end{figure*}

Numerical simulations were initially conducted to validate the feasibility of our approach, followed by imaging tests on individuals and objects within the experimental environment. The experimental system used in the tests is shown in Fig.\ref{fig4}. A supercontinuum laser generated an optical pulse with a width of 6 ps and a wavelength of 1550 ± 25 nm. This pulse was expanded and projected onto the target through the transmitting optical path of the optical lens system. The reflected light was collected by the optical lens system and then coupled into a fiber using a collimator and detected by a Superconducting Nanowire Single Photon Detection (SNSPD) device via another fiber, with a time jitter of 50 ps. To generate a temporal histogram, the time-of-flight interval between the laser and the reflected light was recorded using a Time Digital Converter (TDC) known as the Swabian Time Tagger Ultra. Simultaneously, the millimeter-wave radar (model: TI IWR6843) captured the one-dimensional distance image of the scene, providing a distance resolution of 50 cm and a range of 0 m to 7 m. Both optical system and millimeter-wave radar focus on distance measurements, requiring the detection of the flight time of light/microwaves through space. Due to the inherent delays and errors in the optical path and electrical circuitry of the system, an object actually situated 1m away may yield a measured value of 2m or some other distance. Therefore, distance calibration is necessary. Our calibration method involves placing a mirror/metal reflector at a distance of 1m from the sensor (aligned horizontally with the sensor and verified with a high-precision ruler). We then calibrate the peak value of the detected distance at this position to be 1m. Additionally, the depth camera (model: Orbbec Gemini 2) acquired a three-dimensional depth map of the scene with a range of 0 m to 6 m. The measured impulse response function (IRF) of our system was approximately 200 ps. Specific parameters for each device can be found in the supplemental document.

In the imaging experiment, we conducted tests in an open room to simulate simple background imaging conditions, as depicted in Fig.\ref{fig4}(b). The room had dimensions of 6 m × 6 m, and the target was positioned at a distance of more than 3 m to 4 m from the detector. The target's position within the scene was randomized. To ensure synchronization, the single-pixel single-photon time-of-flight measurement system, millimeter-wave radar system, and depth camera were synchronized every 3 seconds, resulting in the collection of 4000 sets of fused temporal histograms and 3D images (1000 sets of humans and 3000 sets of letters). Our method requires less training data and yields high-quality image reconstruction compared to methods that typically require 6000-10,000 data sets for training algorithms. The data sets were randomly divided into training and testing sets in a 9:1 ratio. The imaging results are presented in Fig.\ref{fig14} (additional experimental results can be found in the supplemental document).

We evaluate the quality of the reconstructed images by computing the Structural Similarity Index Measure(SSIM)\cite{ssim} between the reconstructed images and the ground truth images, SSIM is a perceptual image quality assessment method based on human visual characteristics, which can quantify the degree of distortion in an image and is consistent with human perception of image distortion. SSIM is calculated by comparing the luminance, contrast and structural similarity of two images, as shown in Eq.(\ref{e1})(\ref{e2})(\ref{e3})(\ref{e4})

\begin{equation}
	\label{e1}
	l\left ( x,y \right ) = \frac{2\mu_{x}\mu_{y}+c_{1} }{\mu_{x}^{2}+\mu_{y}^{2}+c_{1}}
\end{equation}

\begin{equation}
	\label{e2}
	c\left ( x,y \right ) = \frac{2\sigma_{x}\sigma_{y}+c_{2} }{\sigma_{x}^{2}+\sigma_{y}^{2}+c_{2}} 
\end{equation}

\begin{equation}
	\label{e3}
	s\left ( x,y \right ) = \frac{\sigma_{xy}+c_{3} }{\sigma_{x}\sigma_{y}+c_{3}} 
\end{equation}

\begin{equation}
	\label{e4}
SSIM\left ( x,y \right ) = \left [ l\left ( x,y \right )^{\alpha } \cdot c\left ( x,y \right )^{\beta } \cdot s\left ( x,y \right )^{\gamma }  \right ] 
\end{equation}

Among them, $ \mu $ is the mean, $ \sigma $ is the variance, $ sigma_{xy} $ is the covariance, the constant $ c_{n}$ is used to avoid division by 0, the determination rule is related to the range of pixel values, and the three power exponents $ \alpha $, $\beta$, and $\gamma$ are used to adjust the importance of the three factors. Generally, the default is 1. At this time, the calculation formula is:

\begin{equation}
	\label{e5}
	SSIM(x, y)=\frac{\left(2 \mu_{x} \mu_{y}+c_{1}\right)\left(2 \sigma_{x y}+c_{2}\right)}{\left(\mu_{x}^{2}+\mu_{y}^{2}+c_{1}\right)\left(\sigma_{x}^{2}+\sigma_{y}^{2}+c_{2}\right)}
\end{equation}

\begin{figure*}[htbp]
	\centering\includegraphics[width=12cm]{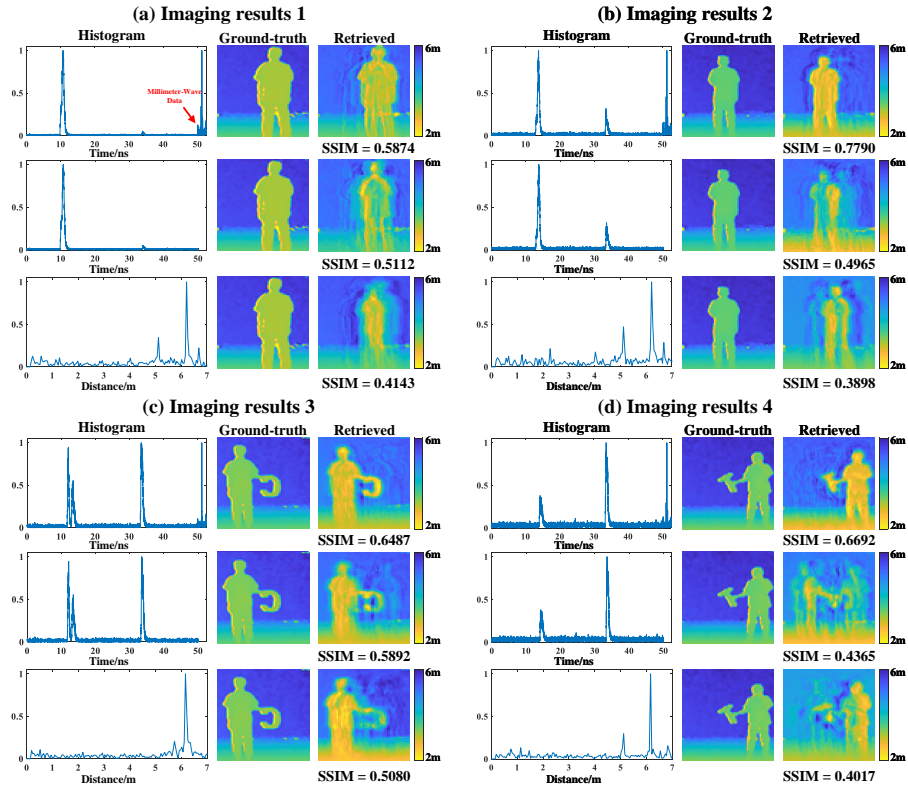}
	\caption{Experiment results. (a)-(d) Imaging result of different people and objects. In each sub-figure, the Histogram, the ground-truth depth maps and the retrieved images are shown from left to right. Moreover, the reconstructed images with fusion data, only single photon data, only millimeter-wave data are shown from top to bottom}
	\label{fig14}
\end{figure*}

In Fig.\ref{fig14}(a)-(d), the first column presents the temporal histograms used for image reconstruction. The second column displays the ground truth depth maps captured by a depth camera for comparison. The third column showcases the reconstructed images based on the histograms from the first column. Each sub-figure depicts the reconstructed images with fusion data, only single-photon data, and only millimeter-wave data, arranged from top to bottom. Among the 400 testing data sets, our proposed fusion method exhibits an average structural similarity index (SSIM) of 0.6576, surpassing the SSIM of the single-photon method (0.6389) and the radar method (0.5266).

In Fig.\ref{fig14}(a) and (b), the fusion method effectively reconstructs clear images of a person. However, the single-photon-only method lacks background information, leading to an imprecise determination of the person's lateral position and resulting in symmetry blur around the person. The radar results reveal that the larger IRF causes the loss of certain details in the shapes, such as incomplete recovery of arms or legs.

In Fig.\ref{fig14}(c), imaging experiments were performed with a person holding the letter "C" (with dimensions of 40 cm × 30 cm). It is evident that the fusion method achieves the highest imaging quality. The single-photon-only method displays an artifact on the right side of the reconstructed image, while the radar method fails to produce a clear image of the letter "C" due to insufficient resolution.

In Fig.\ref{fig14}(d), another imaging experiment was conducted using the letter "T" (with dimensions of 30 cm × 20 cm). It is evident that when the object being imaged is smaller, both the photon-only and radar methods yield poor image reconstruction results, making it challenging to distinguish the person and the letter. However, when the two sets of data are fused together, the object can be reconstructed clearly.

The obtained results demonstrate that the fusion-data-driven 3D imaging method effectively addresses the issue of symmetry blur in single-pixel single-photon 3D imaging. By utilizing a fusion-based approach, the mapping relationship between 3D imaging and time-of-flight histograms is established more effectively, leading to enhanced system performance and robustness.

\section{Discusion}
We incorporated millimeter-wave radar as an auxiliary sensing sensor in our proposed method and experiments to eliminate symmetry blur in imaging and enhance imaging quality. Millimeter-wave radar was selected due to its affordability, all-weather operability, and simplicity of data. These advantages enabled us to achieve considerable improvements in image quality at a relatively low cost.

In the experiment, adhering to the principles of binocular vision, it was necessary to maintain a certain separation distance, referred to as the baseline, between the single-pixel SPD and the millimeter-wave radar. The data obtained from both sensors consisted of 1D temporal data, eliminating the requirement for sensor calibration. It is essential to avoid selecting a baseline that is too close, as this would result in minimal variations in the collected data when objects are in motion. Due to the limited detection field of view of the two sensors, an excessively large baseline distance is also impractical. While a larger baseline enhances imaging accuracy, it introduces blind spots. The imaging performance associated with different baseline distances can be evaluated in subsequent studies.

For data fusion, we adopted a straightforward approach by directly concatenating the 1D single-photon temporal histogram with the millimeter-wave radar data. This simplified the data processing procedure, avoided increasing the complexity of the artificial neural network (ANN), and yielded satisfactory imaging results. More complex and in-depth data fusion methods can further improve the quality of imaging, which is where future work can be improved. The combined utilization of the all-weather operation of millimeter-wave radar and the low-light detection capability of single-photon detectors endowed our proposed method with excellent imaging potential even in extreme conditions such as foggy and rainy weather.

In addition to millimeter-wave radar, we employed a 1550 nm laser as the emission source for the optical system. Compared to the previous work that used 550 nm, the 1550 nm laser offered improved eye safety and enabled higher power output. In current autonomous driving vehicles, lidar and millimeter-wave radar are widely employed independently for detection purposes. However, our proposed method integrates the raw data from both sensors, resulting in a robust imaging system. Consequently, our method holds significant potential for applications in unmanned autonomous navigation platforms and autonomous driving.

During the ANN training, the ground truth collected by the depth camera determines the highest resolution of the system. The IRF of the single-photon system determines the image reconstruction performance of the algorithm. When the IRF is smaller, the reconstructed image is closer to the ground truth.

\section{Conclusion}
Instead of utilizing structured light illumination or laser scanning, single-pixel 3D imaging leverages data-driven imaging retrieval algorithms to convert the 1D temporal histogram obtained from the scene into a 3D depth map. However, the inherent limitations of single-pixel imaging may give rise to issues such as symmetry blur in data-driven image retrieval algorithms. In this paper, we present a novel single-pixel 3D imaging method that integrates temporal data from a single-photon detector and a millimeter-wave radar. Specifically, our approach involves capturing temporal histograms of objects using both a single-pixel single-photon detector and a millimeter-wave radar. The acquired data are then fused and employed to train a neural network based on the MLP algorithm, capable of reconstructing 3D images. To validate the performance of our proposed method, we conducted numerical simulations and experimental measurements. Remarkably, the results demonstrate the superiority of our approach over relying solely on a single-pixel single-photon detector, as it significantly eliminates the impact of symmetrical blur on image reconstruction. Our proposed method exhibits exceptional imaging performance and robustness compared to existing approaches. The combination of single photon detector and millimeter-wave radar enables the development of a novel 3D image reconstruction system, which holds tremendous potential for various applications, including unmanned autonomous navigation platforms, forward-looking imaging in vehicles and indoor security monitoring.

\section*{Acknowledgement}
This work is supported by Shenzhen Science and Technology Program (Grant No. JCYJ2022081 8102014029), National Natural Science Foundation of China (Grant No. 62171458), National key research and development program (Grant No. 2021YFB2802004).


\bibliography{sample}






\end{document}